\documentclass[conference]{IEEEtran}
\IEEEoverridecommandlockouts

\usepackage{amsmath,amssymb,amsfonts, mathtools,mathrsfs}
\newcommand{\colvec}[2][.9]{%
  \scalebox{#1}{%
    \renewcommand{\arraystretch}{1}%
    $\begin{bmatrix}#2\end{bmatrix}$%
  }
 }

\usepackage{graphicx}
\usepackage{xcolor}

\usepackage{hyperref}
\usepackage[normalem]{ulem}
\usepackage{booktabs}
\usepackage{graphicx}
\usepackage{subcaption}

\usepackage[
placement=top,
angle=0,
color=red,
scale=1.1,
hshift=0,
vshift=-15
]{background}
\backgroundsetup{contents=\bf{----------------------------------------------------------------- PREPRINT -----------------------------------------------------------------}}
\begin{document}

\title{RotorSuite: A MATLAB/Simulink Toolbox \\for Tilt Multi-Rotor UAV Modeling
}

\author{\IEEEauthorblockN{Nicola Cigarini$^{1,2}$, Giulia Michieletto$^{1,3}$, Angelo Cenedese$^{1,4}$}
\IEEEauthorblockA{\textit{$^1$Department of Information Engineering} \textit{University of Padova},
Vicenza, Italy\\
\textit{$^2$Department of Land, Environment, Agriculture and Forestry, University of Padova}, Legnaro, Italy\\
\textit{$^3$Department of Management and Engineering}, \textit{University of Padova}, Padova, Italy \\
\textit{$^4$Department of Industrial Engineering}, \textit{University of Padova},
Padova, Italy \\
corresponding contact: \texttt{nicola.cigarini@unipd.it}}
}

\maketitle

\begin{abstract}
In recent years, aerial platforms have evolved from passive flying sensors into versatile, contact-aware robotic systems, leading to rapid advances in platform design. Standard coplanar and collinear quadrotors have been complemented by modern tilted and tilting multi-rotor platforms with enhanced maneuverability. To properly analyze, control, and validate the performance of these emerging platforms, an accurate modeling step is required; however, this can be time-consuming, user-dependent and error-prone. To address this issue, we propose a MATLAB/Simulink toolbox for modeling and simulating the dynamics of a broad class of multi-rotor platforms through both an analytical and physics-based approaches. The toolbox, named \textit{RotorSuite}, is provided with comprehensive documentation and example use cases, representing a valuable tool for didactic, research, and industrial development purposes.
\end{abstract}


\section{Introduction}

Motivated by the growing applications of aerial robotics, conventional coplanar and collinear quadrotors have been rapidly complemented by fully and over-actuated platform designs, characterized by superior maneuverability.  Among these,  both tilted and tilting multi-rotor platforms have become particularly popular due to their high versatility in task execution~\cite{hamandi2023fullpose,zhang2025overactuated}. Recent literature provides numerous examples of platforms with propellers whose spinning axes are arbitrarily oriented, either according to fixed or time-varying cant and/or dihedral angles, which may be identical or different across the rotors~\cite{aboudorra2024modelling}. In parallel, a variety of dynamic modeling approaches and simulation tools for aerial robots are now available for both academic and commercial purposes. While these solutions are often well-constructed and functional, they are typically tailored to specific platforms and offer limited flexibility with respect to design modifications~\cite{idrrissi2022review}.

Leveraging the well-established, stable, and actively maintained MATLAB/Simulink environment, we develop a user-friendly toolbox, called \textit{RotorSuite}, to model and simulate the dynamics of a broad class of multi-rotor platforms through a unified framework that offers multiple customizable design options. Our goal is to simplify a traditionally time-consuming, error-prone, and user-dependent task, providing a robust and practical tool to accurately investigate platform capabilities and efficiently assess their performance under different estimation and control solutions.
\medskip

\noindent\textit{Related works:}
Within the MATLAB/Simulink environment, the {UAV Toolbox}~\cite{UAVToolbox} provides different models and supporting features to be used for multi-rotor and VTOL UAV simulation, including state estimation, control design, and hardware-in-the-loop testing. However, vehicle modeling is typically represented through predefined templates, and altering the underlying structure often requires manual restructuring of the model. Similarly, other Simulink toolboxes such as the {Aerospace Blockset}~\cite{AerospaceToolbox} and the {Robotics System Toolbox}{~\cite{roboticstoolbox}}  offer numerous features for platform control and simulation. However, the {Robotics System Toolbox} is primarily aimed at modeling and controlling ground vehicles, while the {Aerospace Blockset} provides only a few components for modeling aerial platforms, mostly fixed-wing vehicles or collinear multi-rotor platforms with star-shaped propeller layouts. Therefore, existing tools fail to provide a cohesive framework for efficiently defining and simulating the dynamic behavior of arbitrary tilted or tilting multi-rotor platforms only through parameters setting.
\medskip

\noindent\textit{Contributions:}
Implemented entirely in MATLAB/Simulink, the developed \textit{RotorSuite} toolbox addresses the need for a flexible, high-fidelity multi-rotor modeling and simulation framework. It allows users to generate simulation models for a broad class of platforms through a concise, programmatic workflow based on a minimal set of configuration parameters.

\textit{RotorSuite} stands apart from existing multi-rotor toolboxes for several reasons. First, it provides a versatile framework that is not limited to specific geometries or rotor layouts. 
Second, it is cross-platform (Windows, Linux, and macOS) and requires only minimal MATLAB/Simulink add-ons to be installed for full functionality. 
Third, due to its streamlined and intuitive configuration process, it is suitable for both research and educational use, without requiring advanced programming skills or integration with external simulation environments. 
In addition to its high configurability, \textit{RotorSuite} features a dual-layer simulation approach that combines MATLAB/Simulink-based numerical models and Simulink-based physically consistent models exploiting 
fully modular Simscape components. Moreover, the toolbox includes documented examples demonstrating its flexibility, parameterization capabilities, and integration with control algorithms.

The remainder of the paper is organized as follows. Section~\ref{sec:foundations} presents the toolbox theoretical foundations. Section~\ref{sec:RotorSuite_structure} illustrates the toolbox contents, while a use case is discussed in Section~\ref{sec:study_case}. Section~\ref{sec:conclusions} reports final remarks.
\medskip

\noindent\textit{Notation}:
The symbols $\mathbb{R}$ and $\mathbb{N}$ denote the sets of real and natural numbers, respectively, and $\mathbb{R}_{\geq 0}$ denotes the set of nonnegative real numbers. Lowercase italic and bold letters represent scalars and (column) vectors, respectively; e.g., $x \in \mathbb{R}$ and $\mathbf{x} \in \mathbb{R}^n$, with $n \in \mathbb{N}$. Uppercase bold letters denote matrices, e.g., $\mathbf{X} \in \mathbb{R}^{n \times m}$, with $n, m \in \mathbb{N}$, and $\mathbf{I}_n \in \mathbb{R}^{n \times n}$ denotes the $n$-dimensional identity matrix. For a vector $\mathbf{v} \in \mathbb{R}^n$, $\mathrm{diag}(\mathbf{v}) \in \mathbb{R}^{n \times n}$ denotes the corresponding diagonal matrix.
The special orthogonal group $\mathbb{SO}(3) = \{ \mathbf{R} \in \mathbb{R}^{3 \times 3} \mid \mathbf{R}\mathbf{R}^\top = \mathbf{I}_3, \det(\mathbf{R}) = 1 \}$ is used to describe three-dimensional rotations. The matrices $\mathbf{R}_x(\alpha)$, $\mathbf{R}_y(\beta)$, and $\mathbf{R}z(\gamma)$ represent elementary rotations about the $x$, $y$, and $z$ axes by the angles $\alpha$, $\beta$, and $\gamma$, respectively. The operator $[\cdot]\times$ maps a vector in $\mathbb{R}^3$ to its associated $3 \times 3$ skew-symmetric matrix, while the operator $(\cdot)^{\vee}$ denotes the inverse mapping.

\begin{table*}[t!]
    \centering
        \caption{Summary of possible modeling multi-rotor platforms in terms of propeller specifications.}
    \label{tab:platforms}
    \begin{tabular}{ccc|ccccc|c}
    \toprule
\multicolumn{3}{c|}{\textbf{position}} & \multicolumn{5}{c|}{\textbf{orientation}} & \textbf{number} \\ \midrule
coplanar(C) & OR & not coplanar (nC) & zero-tilt (-) & {OR} & \multicolumn{3}{c|}{not zero-tilt (T)} & $n$-rotor ($n$R) \\\cline{6-8}
 & AND &  & & & tilted (Ted) & OR & tilting (Ting) &  \\ 
star-shaped (S) & OR & not star-shaped (nS) & & & cant ($\alpha_\bullet$)  & AND/OR & dihedral ($\beta_\bullet)$  & \\
 &  &  & & & interdependent ($\bullet=0$) & OR & independent ($\bullet=i)$  &\\
                   \bottomrule
    \end{tabular} 
\end{table*}

\section{Toolbox Theoretical Foundations}
\label{sec:foundations}

{The toolbox \textit{RotorSuite} is designed to enable the modeling and simulation of the dynamics of a broad class of multi-rotors. More in detail, we account for the platforms characterized by the rotor configuration features listed in Table~\ref{tab:platforms} and clarified in the rest of the section.} 
Although actuation features can vary significantly depending on the propeller arrangement, the dynamic equations of all the resulting platforms are based on the same theoretical foundations: each can be treated as a rigid body acting in 3D space and actuated by its propellers.

Without assuming a specific propeller layout and considering a generic $n$-rotor platform with $n \geq 3$, we introduce the inertial global frame $\mathscr{F}_W = \{ O_W,\; (\mathbf{x}_W,\, \mathbf{y}_W,\, \mathbf{z}_W) \}$ (\textit{world frame}), whose axes are aligned with the canonical basis vectors  $\mathbf{e}_1$, $\mathbf{e}_2$ and $\mathbf{e}_3$ of $\mathbb{R}^3$ for simplicity, and the platform-fixed frame $\mathscr{F}_B = \{ O_B,\; (\mathbf{x}_B,\, \mathbf{y}_B,\, \mathbf{z}_B) \}$ (\textit{body frame}), whose origin is located at the vehicle center of mass. The position and orientation of the $n$-rotor in the world frame are described by the vector $\mathbf{p} \in \mathbb{R}^3$ and the rotation matrix $\mathbf{R} \in \mathbb{SO}(3)$, representing, respectively, the position of $O_B$ in $\mathscr{F}_W$ and the orientation of $\mathscr{F}_B$ with respect to $\mathscr{F}_W$.

The kinematics of the platform is governed by
\begin{align}  
    \label{eq:lin_vel_model}
    \dot{\mathbf{p}} &=\mathbf{v}\\
    \label{eq:ang_vel_model}
    \dot{\mathbf{R}} &= \mathbf{R}\left[\boldsymbol{\omega}\right]_\times
\end{align}
where $\mathbf{v} \in \mathbb{R}^3$ denotes the linear velocity expressed in $\mathscr{F}_W$, and $\boldsymbol{\omega} \in \mathbb{R}^3$ denotes the angular velocity expressed in $\mathscr{F}_B$.

\begin{figure}[t!]
    \centering
    \includegraphics[width=0.65\linewidth,trim={1.45cm 1.1cm 0.75cm 0.65cm},clip]{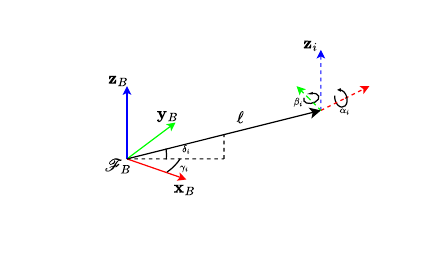}
    \caption{$i$-th propeller position and orientation with respect to the body frame described by parameters $(\ell,\alpha_i,\beta_i,\gamma_i,\delta_i)$.}
    \label{fig:propeller_geometry}
\end{figure}

Any $n$-rotor platform is actuated through the controllable spinning action of its propellers. The position of the $i$-th propeller spinning center, with $i \in {1, \ldots, n}$, is fixed in $\mathscr{F}_B$ and described by the vector $\mathbf{p}_i \in \mathbb{R}^3$ as
\begin{align}
\label{eq:prop_position}
\mathbf{p}_i &= \ell \,
\mathbf{R}_z \left(\gamma_i \right) \mathbf{R}_y \left(\delta_i \right) \mathbf{x}_B, 
\end{align}
where $\ell \in \mathbb{R}_{> 0}$ is the distance between $O_B$ and the propeller center, i.e., the approximated length of the platform arms. In~\eqref{eq:prop_position}, the  \textit{azimuth} angles $\{\gamma_i \in \left[0, 2\pi \right)\}$ define the distribution of the propellers around the $\mathbf{z}_B$ axis, while the \textit{elevation} angles $\{\delta_i \in \left[0, \frac{\pi}{2}\right)\}$  describe their elevation with respect to the $(\mathbf{x}_B,\mathbf{y}_B)$ plane (Figure~\ref{fig:propeller_geometry}). 
For coplanar platforms, $\delta_i = 0$ for all $i \in \{1 \ldots n\}$; for a star-shaped layout, $\gamma_i = (i-1)\frac{2\pi}{n}$, $i \in \{1 \ldots n\}$, with the first propeller along $\mathbf{x}_B$ axis.

The orientation in $\mathscr{F}_B$ of the spinning axis of any $i$-th propeller is identified by the unit vector $\mathbf{z}_{i}\in \mathbb{R}^3$. This vector may vary over time in the case of tilting platforms, whereas it is fixed in the body frame for zero-tilt (collinear) and tilted $n$-rotors. In both cases, we assume that
\begin{equation}
\label{eq:orientations_alphabeta}
\mathbf{z}_{i} =\mathbf{R}_z  \left(\gamma_i \right) \mathbf{R}(\beta_i)\mathbf{R}_x(\alpha_i)\mathbf{z}_B,
\end{equation}
where the \textit{cant} angles $\{\alpha_i \in \left(-\frac{\pi}{2}, \frac{\pi}{2}\right)\}$ describe the spinning axes rotations about the $\mathbf{x}_B$ axis, i.e., along the arm direction, the \textit{dihedral} angles $\{\beta_i \in \left[0, \frac{\pi}{2}\right)\}$ describe rotations about the $\mathbf{y}_B$ axis, i.e., along the direction orthogonal to the arm (Figure~\ref{fig:propeller_geometry}).
Remarkably, for zero tilt-platform $\alpha_i=\beta_i=0$ for all $i \in \{1 \ldots n\}$. In Table~\ref{tab:platforms}, we further distinguish the case of interdependent cant and/or dihedral angles, defined for all rotors through a common parameter $\alpha_0$ 
and/or $\beta_0$.

By rotating about its axis, the $i$-th propeller produces at its center, a \textit{thrust force} $\mathbf{f}_i \in \mathbb{R}^3$ and a \textit{drag moment} $\boldsymbol{\tau}_i^d \in \mathbb{R}^3$. 
Together with the resulting \textit{thrust moment} $\boldsymbol{\tau}_i^t = \mathbf{p}_i \times \mathbf{f}_i \in \mathbb{R}^3$, these depend on the propeller spinning rate $\omega_i \in \mathbb{R}_{\geq 0}$, and are expressed in the body frame as
\begin{align}
\mathbf{f}_i  &=    c_{f_i} \omega_i^2\mathbf{z}_{i},
\label{eq:thrust_force}\\
 {\boldsymbol\tau}_i^d  &= \kappa_i c_{\tau_i} \omega_i^2\mathbf{z}_{i} ,
\label{eq:drag_moment}\\
\boldsymbol{\tau}_i^t  &=  c_{f_i} \omega_i^2  (\mathbf{p}_i \times \mathbf{z}_{i})  \label{eq:thrust_moment}
\end{align}
where  $c_{f_i},c_{\tau_i} \in \mathbb{R}_{\geq 0}$ are constant parameters depending on the rotor geometric features and $\kappa_i \in \{-1,1\}$ allows for distinguishing whether   the $i$-th propeller spins counterclockwise (CCW, $\kappa_i=-1$) or clockwise (CW, $\kappa_i=1$).

Summing up the contributions~\eqref{eq:thrust_force}- \eqref{eq:thrust_moment} for all the propellers, we obtain the \textit{total control force} $\mathbf{f}_c \in \mathbb{R}^3$ and \textit{total control moment} $\boldsymbol{\tau}_c \in \mathbb{R}^3$,  applied at the platform center of mass and expressed in $\mathscr{F}_B$. 
Considering the assignable \textit{control inputs} $\{u_i = \omega_i^2 \in \mathbb{R}_{\geq 0}\}$, it follows that
\begin{align}
\mathbf{f}_c &= \sum_{i=1}^{n} \mathbf{f}_i 
= \sum_{i=1}^{n} c_{f_i}\, \mathbf{z}_{i}\, u_i,   \label{eq:control_force}\\
\boldsymbol{\tau}_c &= \sum_{i=1}^{n} \big(\boldsymbol{\tau}^t_i + \boldsymbol{\tau}^d_i \big)= \sum_{i=1}^{n} \Big(c_{f_i}\, \mathbf{p}_i \times \mathbf{z}_{i} + \kappa_i\, c_{\tau_i}\, \mathbf{z}_{i} \Big) u_i. \label{eq:control_moment}
\end{align}

Defining the \textit{control input vector} 
$
\mathbf{u} = \colvec{ u_1 \; \cdots \; u_n}^\top \in \mathbb{R}^n$, the expressions~\eqref{eq:control_force} and~\eqref{eq:control_moment} can be written as
\begin{align}
\mathbf{f}_c = \mathbf{F} \mathbf{u} \quad \text{and} \quad
\boldsymbol{\tau}_c = \mathbf{M} \mathbf{u}, \label{eq:control_input_matrices}
\end{align}
where the \textit{control force input matrix} $\mathbf{F} \in \mathbb{R}^{3 \times n}$ and the \textit{control moment input matrix} $\mathbf{M} \in \mathbb{R}^{3 \times n}$ depend on the cant and/or dihedral angles, and are  thus time-varying for tilting platforms.

Neglecting second-order effects, the platform dynamics follow from the Newton–Euler equations:
\begin{align}
    \label{eq:force_dynamical_newton}
    m\ddot{\mathbf{p}} &= -mg\mathbf{e}_3 + \mathbf{RFu}\\
    \label{eq:moment_dynamical_newton}
    \mathbf{J}\dot{\boldsymbol{\omega}} &= -\boldsymbol{\omega}\times\mathbf{J}\boldsymbol{\omega} + \mathbf{Mu}
\end{align}
where $g,m>0$ and $\mathbf{J}\in\mathbb{R}^{3\times 3}$ are, respectively, the gravitational acceleration, the total mass of the platform and its positive definite inertia matrix computed in body frame.

\section{Toolbox Description}
\label{sec:RotorSuite_structure}

From a practical perspective, \textit{RotorSuite} presents two main components: the MATLAB functions and the Simulink blocks, with the latter partially integrated with Simscape. These two counterparts are designed to work cooperatively, although characterized by complementary features.

\subsection{MATLAB Functions}

The toolbox includes functions for defining the propeller poses within the multi-rotor frame and computing the corresponding control input matrices according to~\eqref{eq:control_input_matrices}. Moreover, additional analysis functions allow users to investigate the actuation and force–moment decoupling properties of a platform based on the algebraic characteristics of $\mathbf{F}$ and $\mathbf{M}$, as discussed in~\cite{michieletto2018fundamental}. 
Further utility functions support operations in the Lie algebra $\mathfrak{so}(3)$ and the associated rotation group $\mathbb{SO}(3)$. 
Finally, a set of visualization functions enables the generation of graphical representations of the designed platforms, providing intuitive feedback on their configuration.

\subsection{Simulink Blocks}

The Simulink blocks included in \textit{RotorSuite} are designed to enable a dual-layer simulation approach. On the one hand, users can numerically simulate multi-rotor dynamics~\eqref{eq:force_dynamical_newton}-\eqref{eq:moment_dynamical_newton} using MATLAB functions (\textit{analytical approach}). On the other hand, the toolbox includes a library of modular Simscape Multibody components that enables physically consistent simulations of multi-rotor dynamics (\textit{physics-based approach}).

The analytical approach is based on a single configurable Simulink block, named \texttt{TMR - Tilt MultiRotor}. By accessing its parameters, users can define $n$-rotor platforms with $n \in \{3\dots 8\}$, specifying mass and inertia properties, alongside the propeller spinning-axis orientations according to~\eqref{eq:orientations_alphabeta} and their center positions according to~\eqref{eq:prop_position}, or directly by providing their coordinates in the body frame. 
The block \texttt{TMR - Tilt MultiRotor} also allows users to specify the rotor geometric coefficients $\{c_{f_i}, c_{\tau_i}\}$ and the spinning direction through $\{\kappa_i\}$. In addition, the simulation initial conditions can be defined in terms of platform position and orientation, as well as linear and angular velocities and accelerations. 
Finally, users can select the block inputs, which may consist of either the pair of total control force and moment ($\mathbf{f}_c, \boldsymbol{\tau}_c$) or the control input vector $\mathbf{u}$, together with the cant and/or dihedral angles in the case of tilting platforms. Similarly, the selectable outputs include the platform pose, velocities, accelerations, and the propeller spinning rates $\{ \omega_i\}$.

 The physics-based simulation approach relies on multiple Simscape Multibody blocks to model the main body of the platform, the arms, the motor–propeller assemblies, and the tilting mechanisms. Users can freely assemble these components to approximate real multi-rotor designs in terms of both mechanical structure and mass–inertia distribution.

\subsection{Documentation and Examples}

All elements of the toolbox feature comprehensive documentation aligned with the theoretical foundations outlined in Section~\ref{sec:foundations}. The documentation is directly accessible through the MATLAB help system and is designed to closely follow the standard MATLAB documentation style.

Furthermore, several examples are available to users, including the {MATLAB/Simulink modeling of a conventional coplanar star-shaped zero-tilt quadrotor (\texttt{CS\_4R}, with $n=4$, $\alpha_i=\beta_i=\delta_i = 0$ and $\gamma_i = (i-1)\frac{\pi}{2}$,  $i \in \{1 \ldots 4\}$) and a fully actuated coplanar star-shaped interdependent-cant-tilted hexarotor  (\texttt{CS\_$\alpha_0$-Ted\_6R}, with $n=6$, $\beta_i=\delta_i = 0$, $\gamma_i = (i-1)\frac{\pi}{3}$ and $\alpha_i=f(\alpha_0)$,  $i \in \{1 \ldots 6\}$)}.



\section{Toolbox Use Case}
\label{sec:study_case}

\begin{figure*}[t!]
    \centering
    \begin{minipage}{0.5\textwidth}
        \centering
        \captionof{table}{Use Case Specifications}
        \resizebox{0.5\linewidth}{!}{
        \begin{tabular}{lll}
            \toprule
            $\ell$ & [m] & 0.385\\
            $\delta_i$ & [rad] & 0\\
            $\gamma_i$ & [rad] & $(i-1){\pi}/3$\\
            $\beta_i$ & [rad] & $\beta_0={\pi}/{18}$ \\
            $\alpha_i$ & [rad] & $(-1)^i \cdot \alpha_0, \; \alpha _0 = {5\pi}/{36}$\\\midrule
            ${c_{f,i}}$ & $\left[ \frac{\text{Ns\textsuperscript{2}}}{\text{rad\textsuperscript{2}}} \right]$ & 3.799e-05 \\
            ${c_{\tau,i}}$ & $\left[ \frac{\text{Nms\textsuperscript{2}}}{\text{rad\textsuperscript{2}}} \right]$ &  1.163e-06 \\
            $\kappa_i$ & [] &  $(-1)^i$\\ \midrule
            $m$ & [kg] &  3.500 \\
            $g$ & [m/s\textsuperscript{2}] & 9.810 \\
            $\mathbf{J}^a$ & [kgm\textsuperscript{2}] &  $\text{diag}\left(\colvec{0.155 \; 0.147\; 0.251}\right)$\\
            $\mathbf{J}^p$ & [kgm\textsuperscript{2}] &  $\text{diag}\left(\colvec{0.081\;  0.081\;0.162}\right)$\\
            \bottomrule
        \end{tabular}}
        \label{tab:hexarotor_params}
    \end{minipage}\hfill
    \begin{minipage}{0.5\textwidth}
        \centering
        \includegraphics[width=\linewidth]{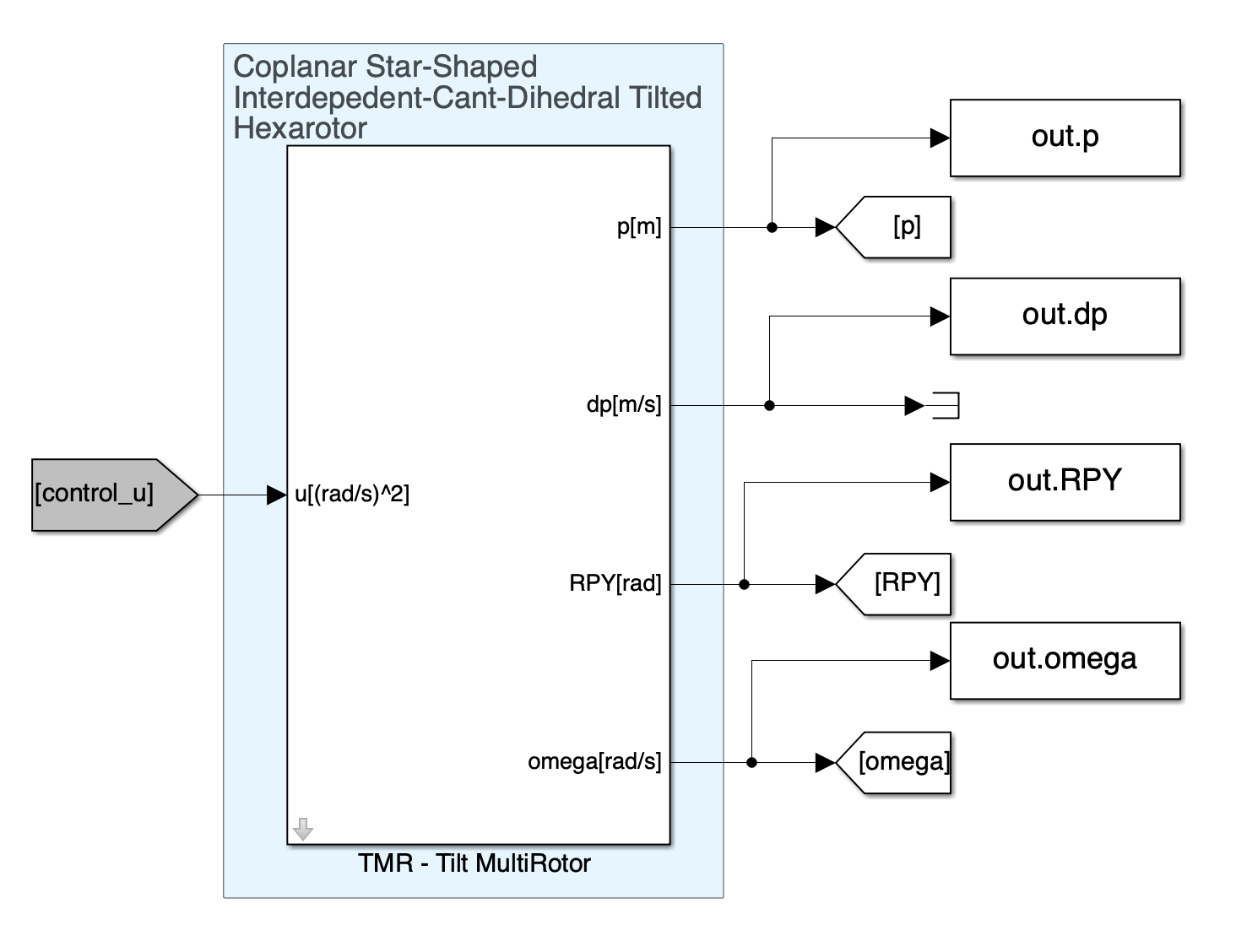}
        \captionof{figure}{Use case analytical simulation setup.}
        \label{fig:numerical_setup}
    \end{minipage}

    \centering
    \includegraphics[width=0.95\linewidth]{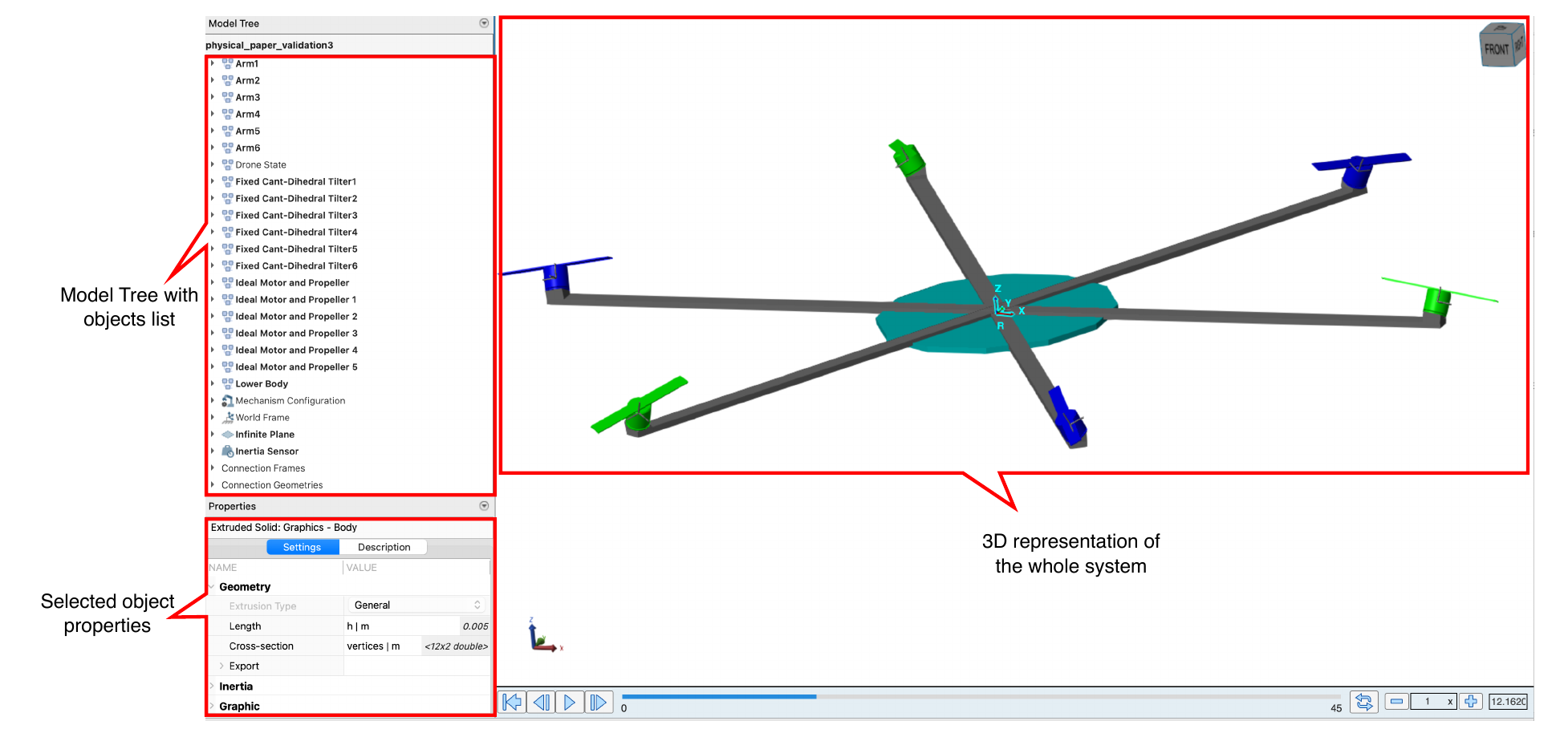}
    \caption{Use case physics-based platform model.}
    \label{fig:simscape_model_view}
\end{figure*}

To demonstrate the versatility, intuitiveness, and effectiveness of the \textit{RotorSuite} toolbox, we present a use case focused on dynamics modeling and validation of a fully actuated coplanar star-shaped interdependent-cant\&dihedral-tilted hexarotor (\texttt{CS\_$\alpha_0\beta_0$-Ted\_6R}), whose features are reported in Table~\ref{tab:hexarotor_params}.

For analytical simulation, the platform is modeled as a rigid body whose inertia is expressed in the body frame via the matrix $\mathbf{J}^a \in \mathbb{R}^{3 \times 3}$ in Table~\ref{tab:hexarotor_params}. This corresponds to a lumped-parameter rigid-body representation of the multi-rotor, implemented via \textit{RotorSuite} by using the Simulink block {\texttt{TMR - Tilt MultiRotor}} as depicted in Figure~\ref{fig:numerical_setup}.

    
    
    
    
\begin{figure*}[t]
   \centering
   \includegraphics[width=0.95\linewidth,trim={0.3cm 0.2cm 0.3cm 0.2cm},clip]{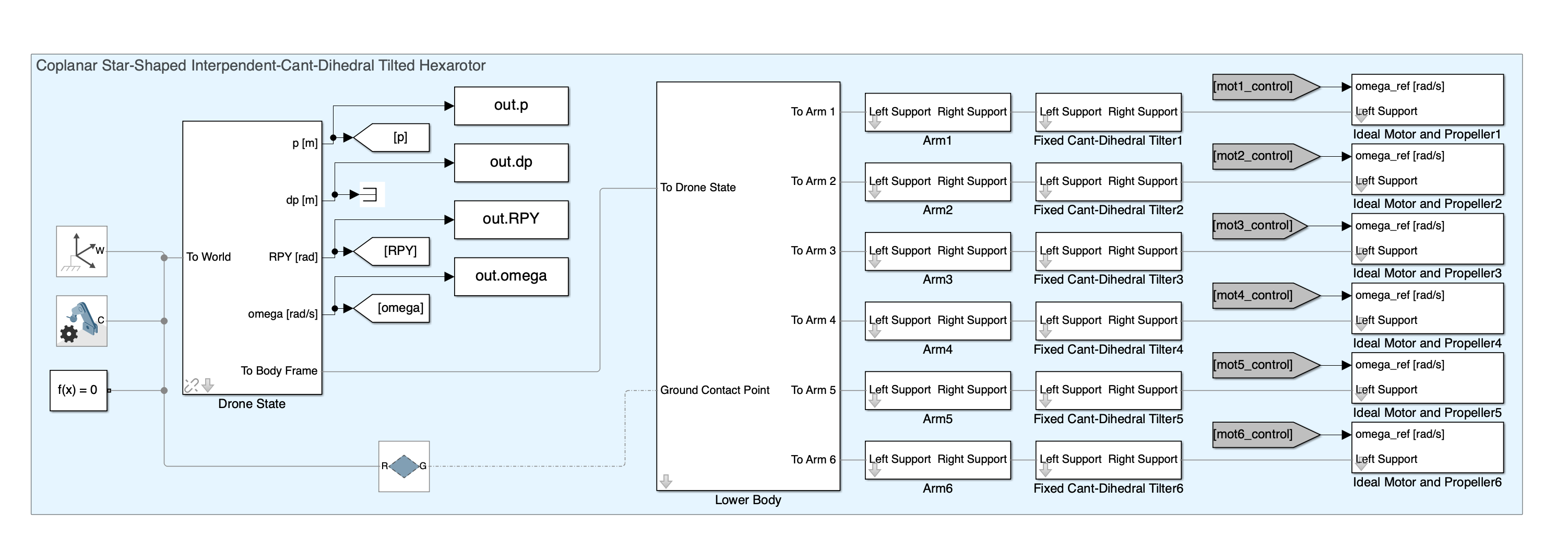}
   \caption{{Use case physics-based simulation setup.}}
   \label{fig:physical_setup}
\end{figure*}

For the physics-based simulation, the platform structure and actuators configuration are explicitly modeled as reported in Figure~\ref{fig:simscape_model_view}, which displays the Simscape model viewer with the full 3D representation, model tree, and parameters of the selected block. Notably, the Simscape model is a higher-fidelity representation with a more detailed mass distribution. Consequently, while the multi-rotor mass remains the same as in the analytical simulation setup, the inertia is evaluated using built-in Simscape blocks. This is indicated as $\mathbf{J}^p$ in Table~\ref{tab:hexarotor_params} and differs from $\mathbf{J}^a$ by a scale factor. Compared to the analytical setup, the physics-based setup, shown in Figure~\ref{fig:physical_setup}, exhibits greater complexity. Nevertheless, thanks to the modular architecture of the Simscape components and the compact parametric description required for their configuration, the increased modeling fidelity does not compromise ease of implementation, confirming the scalability and rapid prototyping capabilities of \textit{RotorSuite}.

\begin{figure*}[]
    \centering
    
    \begin{subfigure}[b]{0.3\textwidth}
        \centering
        \fbox{\includegraphics[width=\textwidth]{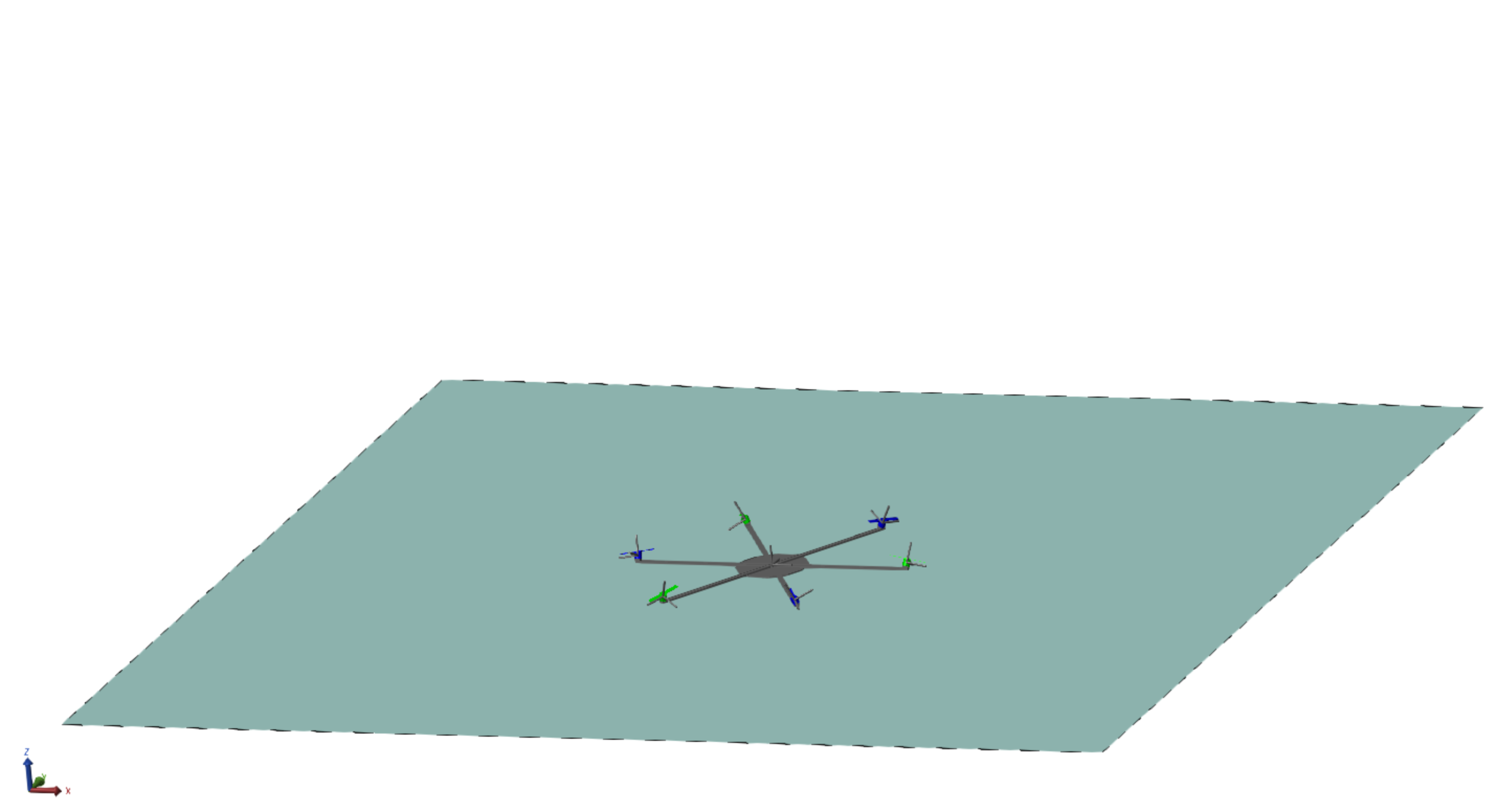}}
        \caption{$t$=0.0s: starting condition}
        \label{fig:snap_traj_1}
    \end{subfigure}
    \hfill
    \begin{subfigure}[b]{0.3\textwidth}
        \centering
        \fbox{\includegraphics[width=\textwidth]{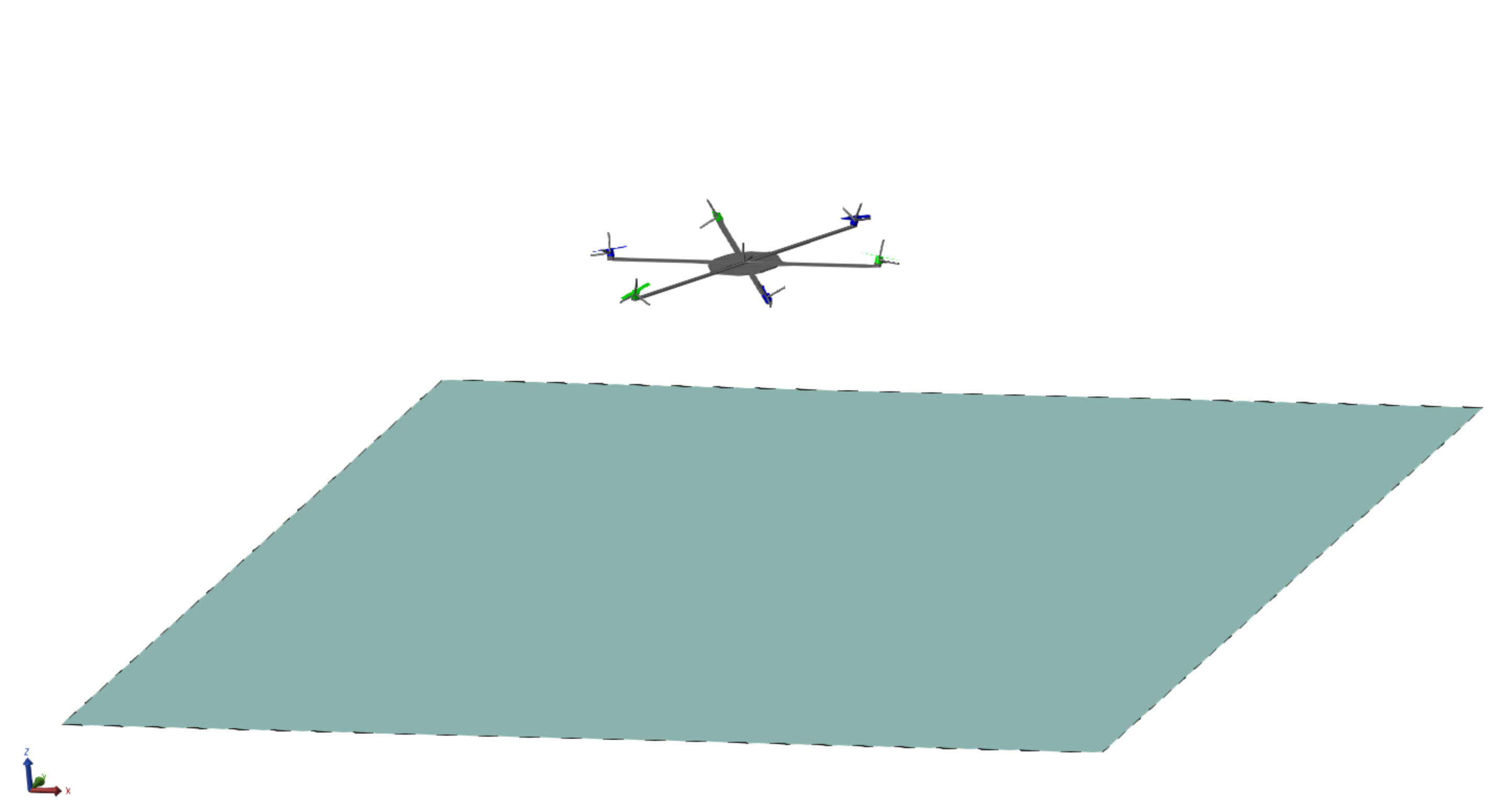}}
        \caption{$t$=8.0s: take off}
        \label{fig:snap_traj_2}
    \end{subfigure}
    \hfill
    \begin{subfigure}[b]{0.3\textwidth}
        \centering
        \fbox{\includegraphics[width=\textwidth]{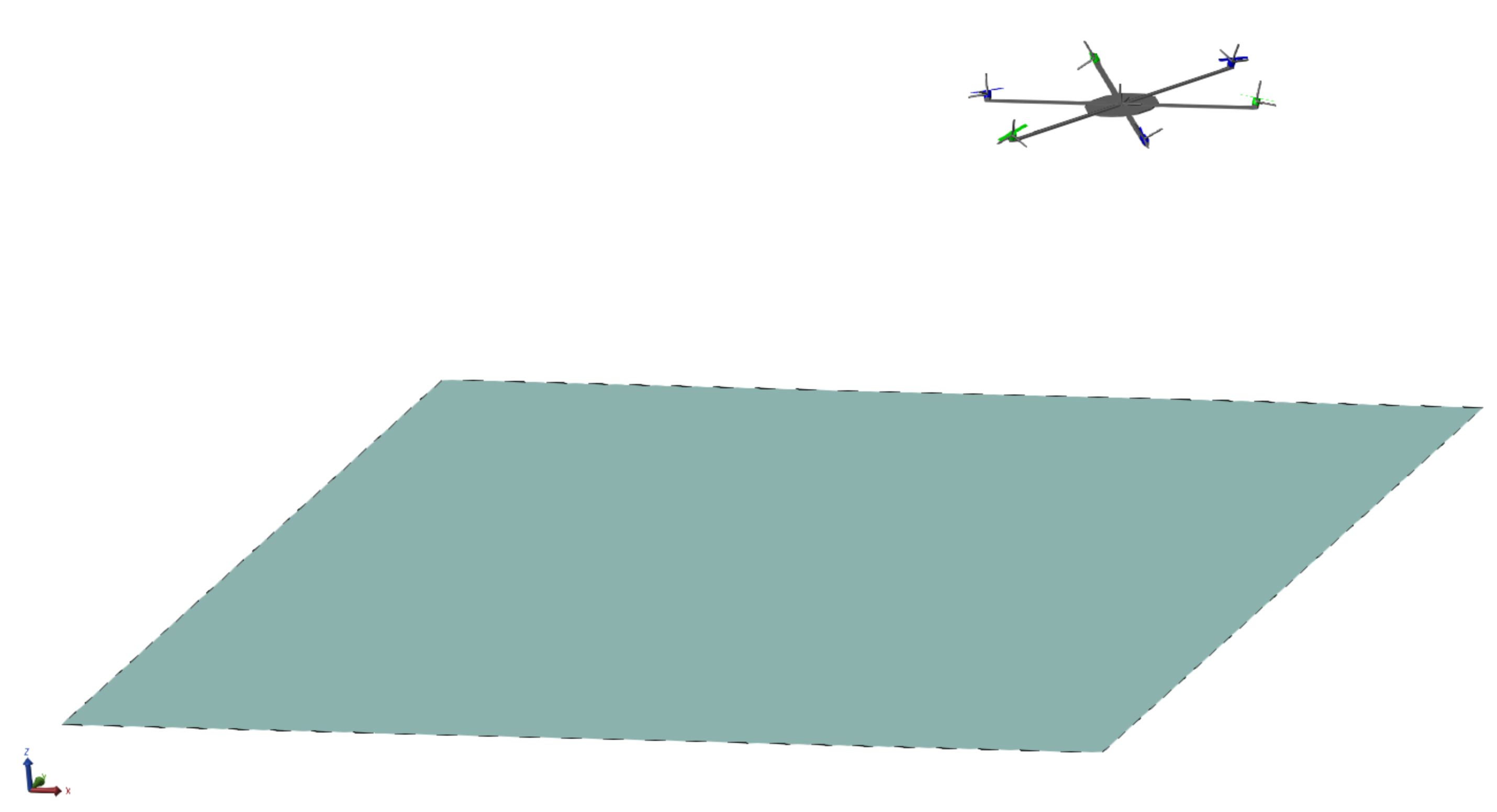}}
        \caption{$t$ = 12.0s: trajectory tracking}
        \label{fig:snap_traj_3}
    \end{subfigure}
    
    \begin{subfigure}[b]{0.3\textwidth}
        \centering
        \fbox{\includegraphics[width=\textwidth]{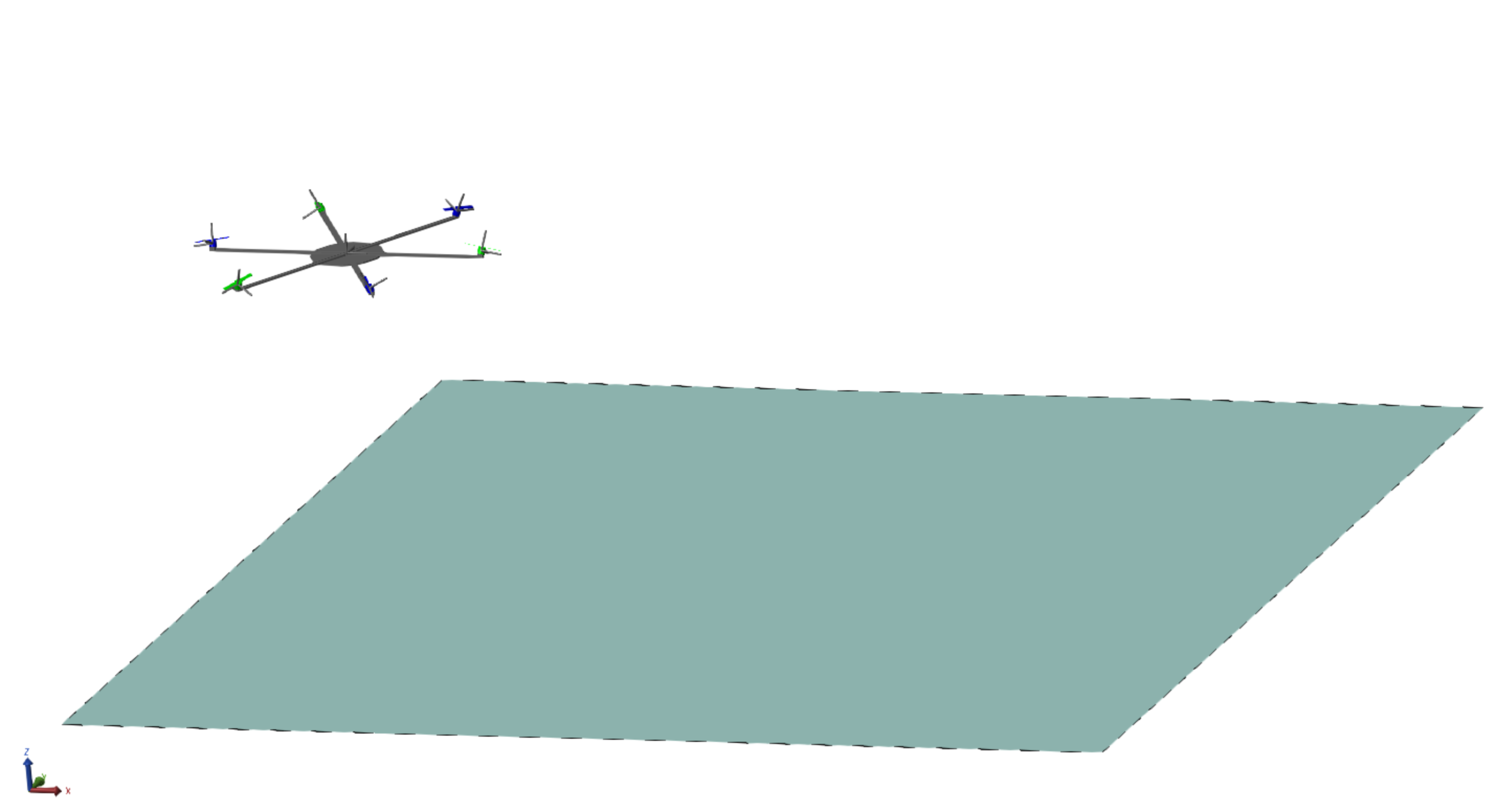}}
        \caption{$t$=30.0s: trajectory tracking}
        \label{fig:snap_traj_4}
    \end{subfigure}
    \hfill
    \begin{subfigure}[b]{0.3\textwidth}
        \centering
        \fbox{\includegraphics[width=\textwidth]{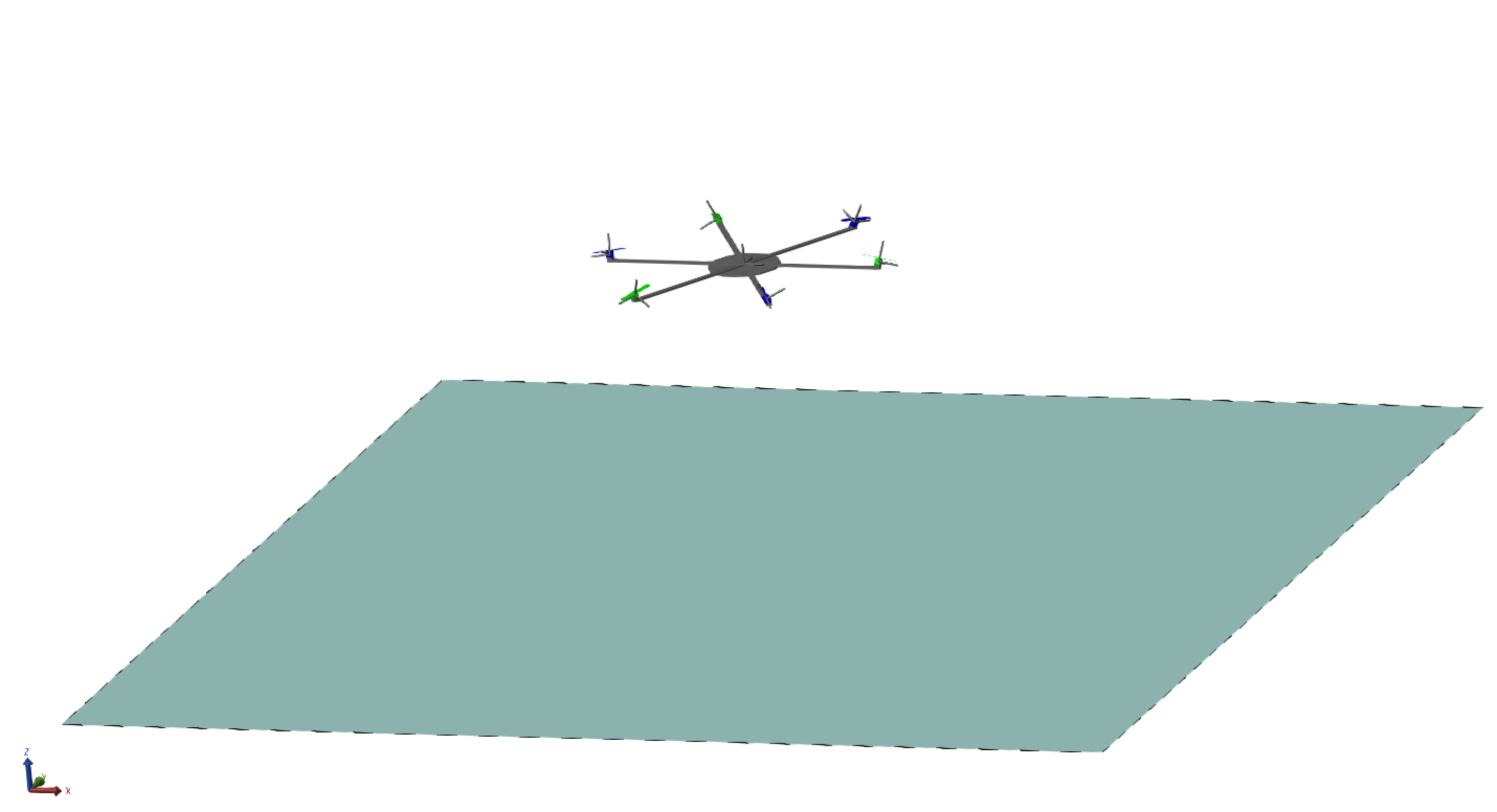}}
        \caption{$t$=37.0s: landing}
        \label{fig:snap_traj_5}
    \end{subfigure}
    \hfill
    \begin{subfigure}[b]{0.3\textwidth}
        \centering
        \fbox{\includegraphics[width=\textwidth]{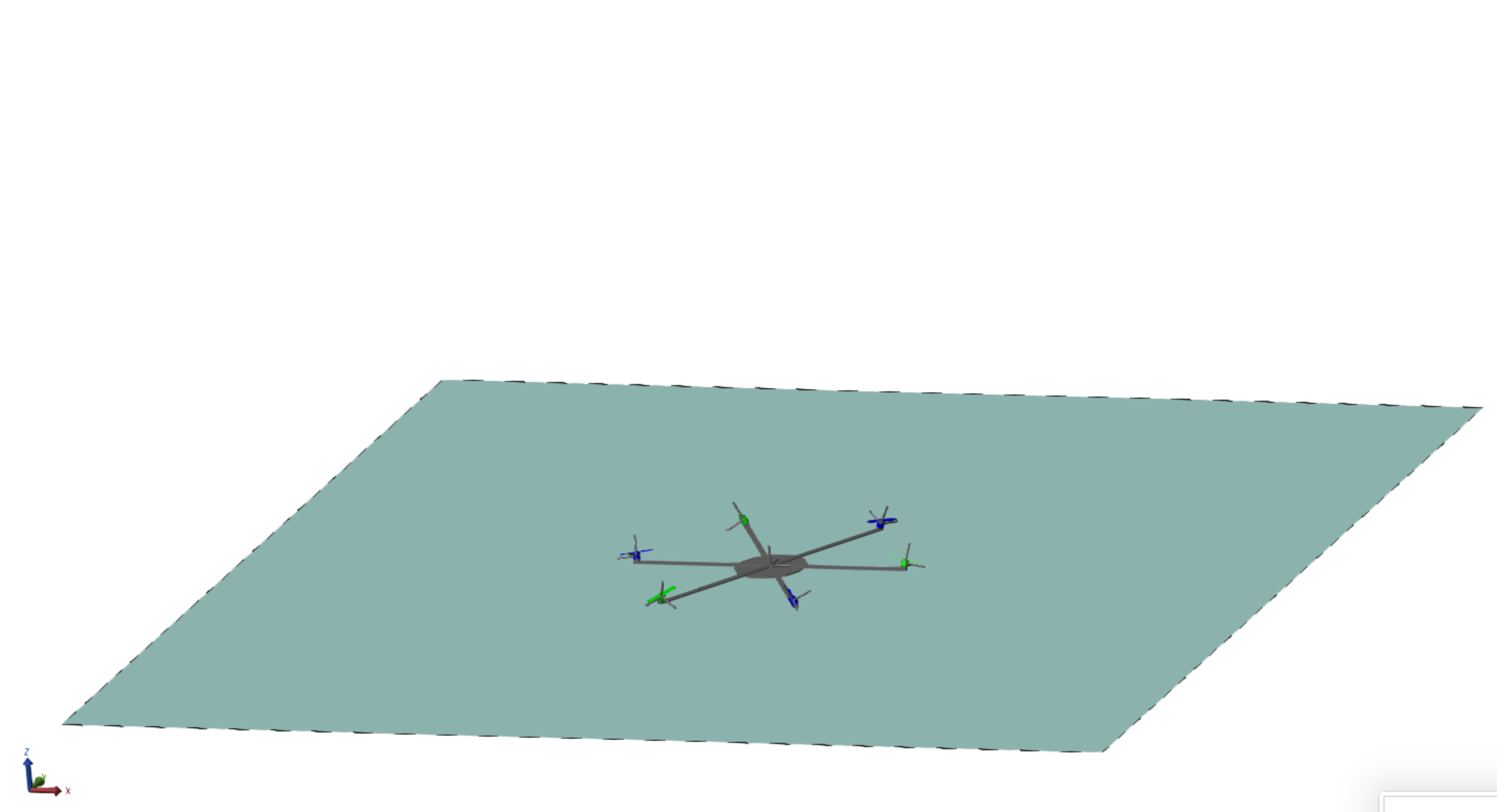}}
        \caption{$t$=45.0s: final condition}
        \label{fig:snap_traj_6}
    \end{subfigure}
    
    \caption{Snapshots of the platform executing the reference maneuver in the physics-based simulation.}
    \label{fig:snap_traj}
\end{figure*}

A geometric controller based on the $\mathbb{SE}(3)$ trajectory regulation solution presented in~{\cite{ryll2021fast}} is introduced in both analytical and physics-based simulation setups, relying on the control force and moment input matrices~\eqref{eq:control_input_matrices}. For the sake of completeness, we report in Table~\ref{tab:controller_gains} the selected controller gains, adopting the notation of~{\cite{ryll2021fast}}.
The Simulink implementation of the controller differs in its outputs between the analytical and physics-based simulation setups: in the former, the control input vector $\mathbf{u}$ feeds the \texttt{TMR- Tilt MultiRotor} block (gray label, Figure~\ref{fig:numerical_setup}); in the latter, the propeller spinning rates $\{\omega_1,\dots,\omega_6\}$ are applied individually to the actuator blocks (gray labels, Figure~\ref{fig:physical_setup}).

\begin{table}[t]
    \centering
    \caption{Controller Gains}
    \label{tab:controller_gains}
    \begin{tabular}{l| l}
        \toprule
         $\begin{aligned}K_{p} &= \text{diag}\left(\colvec{   7  \; 7  \; 60}\right)\\
         K_{p_i}&= \text{diag}\left(\colvec{   3  \; 3  \; 5  }\right)\\
         K_{v} &= \text{diag}\left(\colvec{ 14 \; 14 \; 20}\right)
         \end{aligned}$ &         $\begin{aligned}
             K_{R} &= \text{diag}\left(\colvec{   90 \; 90 \; 90}\right)\\
             K_{R_i} &= \text{diag}\left(\colvec{   25 \; 25 \; 10 }\right)\\
             K_{\omega} &= \text{diag}\left(\colvec{   25 \; 25 \; 5  }\right)
         \end{aligned}$ \\
        \bottomrule
    \end{tabular}
\end{table}

To validate both simulation setups, the platform is tasked to follow a figure-eight trajectory in the $(\mathbf{x}_W,\mathbf{y}_W)$ plane with constant altitude and fixed orientation, defined in terms of reference position $\mathbf{p}_r \in \mathbb{R}^3$ and orientation $\mathbf{R}_r \in \mathbb{SO}(3)$ as 
\begin{align}
\mathbf{p}_r = \colvec{ \sin\left(\frac{\pi}{15} t\right) & \sin\left(2\frac{\pi}{15}\right) & 1}^\top \quad \text{and} \quad \mathbf{R}_r = \mathbf{I}_3
\end{align}
The reference maneuver includes takeoff and landing, each lasting 8~s. Figure~\ref{fig:snap_traj} illustrates the different maneuver phases through selected snapshots of the physics-based simulation, captured directly from the Simscape model viewer.

\begin{figure}[t!]
    \centering
    \includegraphics[width=0.98\linewidth,trim={0cm 18cm 0cm 0cm},clip]{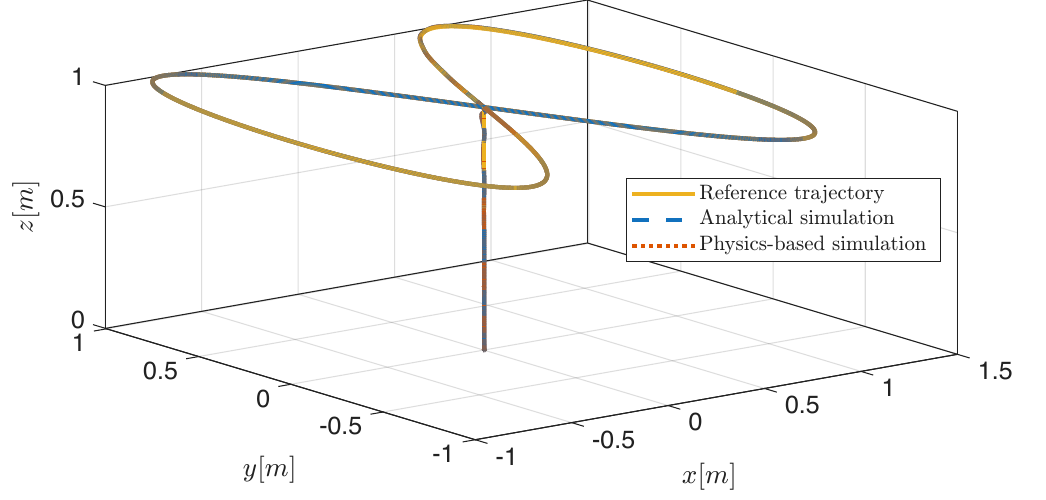}
    \caption{Analytical VS. physics-based simulation tracking performance: 3D position trajectory.}
    \label{fig:traj_pos_att}
    \includegraphics[width=0.98\linewidth,trim={0cm 0cm 0cm 8.5cm},clip]{images/traj_pos_att_err.pdf}
    \caption{Analytical VS. physics-based simulation tracking performance: position $\mathbf{e}_p^\diamond$ (top panel) and orientation $\mathbf{e}_R^\diamond$   (bottom panel) errors.}
    \label{fig:traj_pos_att_err}
\end{figure}

Figure~\ref{fig:traj_pos_att} shows that the closed-loop dynamics is very similar when adopting the two simulation approaches. 
For a better insight, Figure~\ref{fig:traj_pos_att_err} reports the trend of the position and orientation errors, $\mathbf{e}_p^{\diamond}, \mathbf{e}_R^{\diamond} \in \mathbb{R}^3$, computed as 
\begin{align}
    \mathbf{e}_p^{\diamond} &=\colvec{\mathbf{e}_{p,1}^{\diamond} &\mathbf{e}_{p,2}^{\diamond} & \mathbf{e}_{p,3}^{\diamond}}^\top= \mathbf{p}_r - \mathbf{p}^{\diamond}\\
    \mathbf{e}_R^{\diamond} &= \colvec{\mathbf{e}_{R,1}^{\diamond} &\mathbf{e}_{R,2}^{\diamond} & \mathbf{e}_{R,3}^{\diamond}}^\top =\frac{1}{2}\left(\mathbf{R}_r - \mathbf{R}^{\diamond}\right)^\vee
\end{align}
{where ${\diamond} = \{\mathrm{a},\mathrm{p}\}$} to distinguish between analytical and physics-based simulations.
For the position errors, no noticeable discrepancies are observed between the performances obtained with the two setups. In both cases, the deviations of the $\mathbf{e}_{p,1}^{\diamond}$ and $\mathbf{e}_{p,2}^{\diamond}$ components from zero remain below 10 cm and occur during the transients between maneuver phases.
Regarding the orientation errors, in the physics-based simulation setup the trends exhibit limited but perceptible deviations from zero, which do not appear in the analytical simulation setup. This discrepancy can be attributed to the difference between the inertia matrices $\mathbf{J}^a$ and $\mathbf{J}^p$. For quantitative validation, Table~\ref{tab:mean_errors} reports the mean absolute tracking error over the entire simulation in both analytical and physics-based setups.

\begin{table}[t!]
\centering
\caption{Mean Absolute Tracking Errors}
\label{tab:mean_errors}
\begin{tabular}{lccc}
\toprule
& $\diamond = a$ &$\diamond = p$ & difference \\
\midrule
$\mathbf{e}_{p,1}^{\diamond}$ [m]        &  1.7216e-03  &  1.7089e-03  &  1.2671e-05 \\
$\mathbf{e}_{p,2}^{\diamond}$ [m]        &  3.4434e-03 &   3.4200e-03  &  2.3341e-05 \\
$\mathbf{e}_{p,3}^{\diamond}$ [m]        &  1.5840e-05  &  5.667e-05  &  4.0830e-05 \\
$\mathbf{e}_{R,1}^{\diamond}$ [rad]  &  8.5025e-18  &  3.3220e-04  &  3.3220e-04  \\
$\mathbf{e}_{R,2}^{\diamond}$ [rad]  &  7.0063e-18  &  1.0891e-04  &  1.0891e-04  \\
$\mathbf{e}_{R,3}^{\diamond}$ [rad]  &  7.1449e-18  &  1.6628e-07  &  1.6228e-07  \\
\bottomrule
\end{tabular}
\end{table}

\begin{figure}[t!]
     \centering
\includegraphics[width=0.98\linewidth,trim={0cm 0cm 0cm 0.1cm},clip]{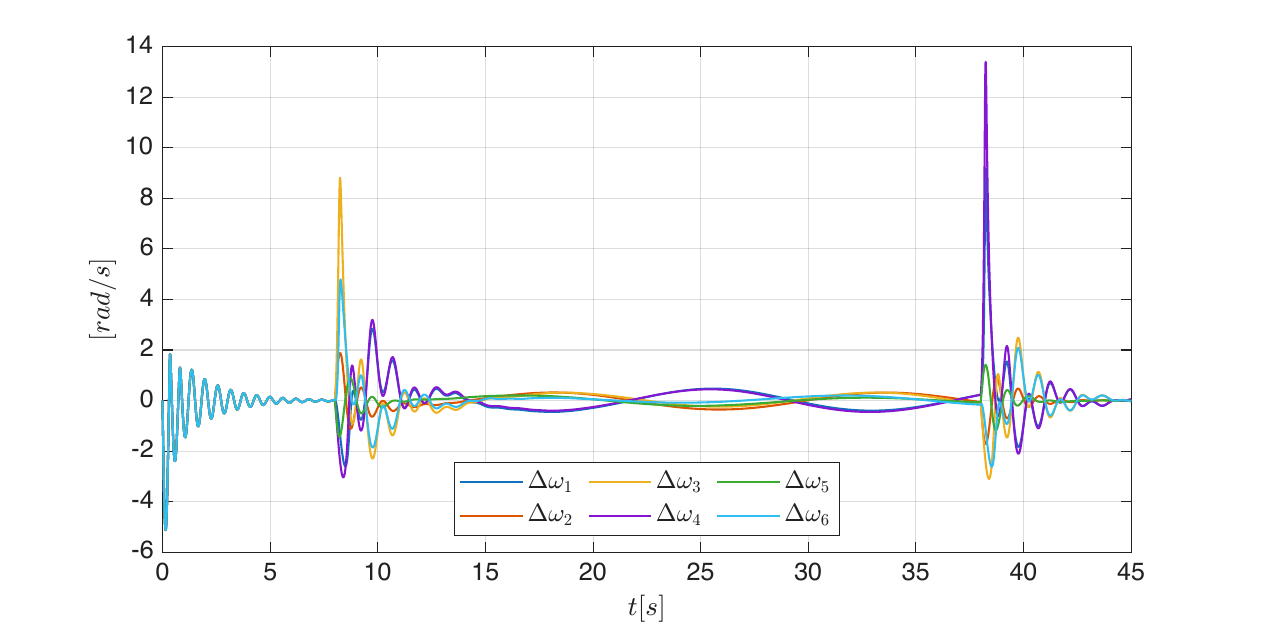}
    \caption{Analytical VS. physics-based simulation tracking performance: propellers spinning rate difference $\Delta\omega_i$.}
    \label{fig:motor_errors}
\end{figure}

To further analyze the consistency between the two simulation approaches, Figure~\ref{fig:motor_errors} shows the differences between the propeller spinning rates in the physics-based and analytical setup, defined as $\Delta \omega_i = \omega_i^{p} - \omega_i^{a} \in \mathbb{R}$ and expressed in rad/s. The largest discrepancies occur in correspondence to the switch among the reference maneuver phases, wherein the errors trends also differ. Indeed, in the analytical simulation setup, propeller spinning rates are applied instantaneously to the dynamics model of the platform. In the physics-based simulation setup, the inclusion of actuator dynamics introduces a finite rise time before steady-state conditions are reached. Nonetheless, despite these transients, the trend of the signals $\Delta \omega_i$ is bounded in $[-2,2]$rad/s. These results indicate that the additional actuator modeling does not significantly affect steady-state tracking, further confirming the consistency between the two developed simulation approaches.


\section{Conclusions}
\label{sec:conclusions}
This paper presents \textit{RotorSuite}, a novel MATLAB/Simulink toolbox for modeling and simulating the dynamics of zero/no-zero tilt multi-rotors through both analytical and physics-based approaches. Extensive tests on different platforms demonstrate both the intuitiveness and versatility of the designed toolbox, and the strong consistency between the two developed simulation approaches. On the one hand, \textit{RotorSuite} allows users to account for multiple propeller layouts through a suitable parameters setting. On the other hand, although the physics-based setup captures additional effects, such as detailed inertia modeling and actuator dynamics, its steady-state tracking performance remains closely aligned with the analytical counterpart, with verified discrepancies limited to transient phases. 


\bibliographystyle{IEEEtran}
\bibliography{IEEE_biblio}

\end{document}